\documentclass{article}
\usepackage{packages}
\newcommand{\zhihao}[1]{{[\textbf{\color{blue}  Zhihao:{#1}}]}}

\title{Token-based Spatiotemporal Representation of the Events}
\name{Bin Jiang\footnotemark[1]$^{1,\dag}$, Zhihao Li\footnotemark[1]$^{1,\dag}$, M. Salman Asif$^\ddag$, Xun Cao$^\dag$, and Zhan Ma$^{\star\dag}$}
\address{$^\dag$Nanjing University, $^\ddag$University of California, Riverside}

\begin{document}
%
\maketitle

\footnotetext[1]{These authors contributed equally. $^\star$Z. Ma is the corresponding author (mazhan@nju.edu.cn).}

\begin{abstract}
The event camera's low power consumption and ability to capture microsecond brightness changes make it attractive for various computer vision tasks. Existing event representation methods typically convert events into frames, voxel grids, or spikes for deep neural networks (DNNs). However, these approaches often sacrifice temporal granularity or require specialized devices for processing. This work introduces a novel token-based event representation, where each event is considered a fundamental processing unit termed an event-token. This approach preserves the sequence's intricate spatiotemporal attributes at the event level. Moreover, we propose a Three-way Attention mechanism in the Event Transformer Block (ETB) to collaboratively construct temporal and spatial correlations between events. We compare our proposed token-based event representation extensively with other prevalent methods for object classification and optical flow estimation. The experimental results showcase its competitive performance while demanding minimal computational resources on standard devices.
\end{abstract}
\section{Introduction}
\label{sec:intro}
Unlike conventional cameras, which produce standard intensity frames at fixed intervals, the event camera emulates the biological behavior of the retina~\cite{survey_2020}. It captures the brightness change of a pixel asynchronously within microseconds, triggered whenever it surpasses a preset threshold. This process results in a sequence of N-event formally denoted as $\mathbf{E}=\{{\boldsymbol{e}_{i}}\}_{i=1}^{N}$. Each event $\boldsymbol{e}_i$ is a tuple that records pixel position $\vec{u}_i = (x_i, y_i)$, where the event is located, polarity $p_i \in \{-1, 1\}$ and timestamp $t_i$. $p_i$ indicates the increase or decrease of pixel brightness. To effectively utilize this asynchronous, discrete, non-uniformly distributed data in DNNs, the event sequence should be represented appropriately, such as {\it frame-based}, {\it voxel-grid-based}, and {\it spike-based} representations. According to the characterization principle, frame-based representation~\cite{hots, nCars, EST, AMAE, M-LSTM, MVF-Net,evt,dvsvit, ev_flownet, firenet} loses temporal information as it discretizes events into fixed time intervals, leading to the loss of precise temporal order and duration between events. Voxel-grid-based representation~\cite{Eventnet, asl_dvs, EV-VGCNN, VMV-GCN,TORE} groups events into voxels over short time periods, but this can sacrifice fine temporal detail. Furthermore, smaller intervals improve detail but increase computation and memory needs. Spike-based representation~\cite{gabor_snn, PLIF, NDA, VPT-STS, spike-evflownet} necessitates specialized algorithms and devices to handle the asynchronous nature of spike events effectively, as conventional continuous data processing methods are not directly applicable to this type of representation. 

In this paper, we take a different approach by treating each individual event in the sequence as a token and converting these event-tokens into fixed-dimensional embedded vectors event by event that preserve their spatiotemporal characteristics. To address the limitations of independent embedded vectors in capturing the internal spatiotemporal variations of event sequence, we introduce an Event Transformer Block that consists of a three-way attention mechanism: Local Attention (LXAttn) operates over a one-dimensional time span to capture local spatiotemporal correlations, Sparse Convolution Attention (SCAttn) performs two-dimensional spatial encoding to enhance the local spatial relationships between events, and Global Attention (GXAttn) aggregates spatiotemporal details from a downscale sequence in a holistic manner. Different from spike-based representation, our model can be easily implemented on mainstream devices since we only use common operators like multilayer perceptron (MLP) and sparse convolution.

Unlike other transformer-based approaches~\cite{evt,dvsvit}, where event image patches are treated as fundamental processing units, leading to an essentially frame-based event representation that discards the temporal information of the sequence, our approach handles sequence event by event, resulting in an event-level spatiotemporal fine-grained representation.


\section{Methodology}

\subsection{Spatiotemporal Embedding}
Essentially, the event sequence consists of individual discrete events, and a new event is triggered when the change in
brightness $L$ of a pixel reaches a threshold $C$: $\Delta L({{{\vec{u}}}_{i}},{{t}_{i}})=L({{{\vec{u}}}_{i}},{{t}_{i}})-L({{{\vec{u}}}_{i}},{{t}_{i-1}}) = {{p}_{i}}C,{{p}_{i}}\in \{-1,1\}$. This indicates a comprehensive representation of the event sequence necessitates considering the spatiotemporal information and polarity of each event $\boldsymbol e_i=\{\vec{u}_{i},t_i,p_i\}$. Therefore, we treat each individual event as the fundamental processing unit, termed as an event-token, and design a separate spatial embedding layer and temporal embedding layer to collaboratively construct spatiotemporal embedded vectors $\mathbf{F}$. 

Since events are often triggered by object motion within microseconds, the spatial features of individual events are closely related to the distribution of other events in the surrounding pixels. Therefore, we employ a convolutional approach for spatial embedding. As shown in Fig.~\ref{fig:figure_main}(b), we stack all $N$ events into a two-dimensional event image with two feature channels. One channel counts the number of positive polarity events at each pixel position, and the other counts the negative ones. Subsequently, considering the sparsity of events, we employ sparse convolution $\rm SConv()$~\cite{sparseconv} to extract spatial features for events at pixel position $\mathbf{u}=\{{\vec {u}_{i}}\}_{i=1}^{N}$. Sparse convolution is only executed when a pixel is at the center of the convolution kernel. Events at the same pixel position share the same convolution results. Thus, the spatial embedded vectors ${\bold F}_{xyp}=\rm SConv(\mathbf{u},\{ \emph{p}_{i}\}_{i=1}^{N})\in \mathbb{R}^{N \times \textit{C}_{xyp}}$, where $\{p_i\}_{i=1}^{N}$ denote the polarities of all events used for the event count of the feature channel, and $\textit{C}_{xyp}$ donates output channel. 

Although ${\bold F}_{xyp}$ embeds the spatial information of each event, the event image still discards the temporal information of the sequence. \label{sec:TE}Therefore, as shown in Fig.~\ref{fig:figure_main}(c), we further embed the one dimension temporal information of all events into the temporal embedded vectors ${{\mathbf{F}}_{tp}}=\mathbf{p}_{diag}^{+}\cdot (\mathbf{t}\cdot {{\mathbf{W}}^{+}}+{{\mathbf{b}}^{+}})+\mathbf{p}_{diag}^{-}\cdot (\mathbf{t}\cdot {{\mathbf{W}}^{-}}+{{\mathbf{b}}^{-}})\in {{\mathbb{R}}^{N\times {{C}_{tp}}}}$, where $\mathbf{p}_{diag}^{+},\mathbf{p}_{diag}^{-}\in \mathbb{R}^{N \times \textit{N}}$ are diagonal matrices, with $\mathbf{p}_{diag}^{+}(i,i)$ equal to 1 for positive $p_i$, and 0 otherwise; and $\mathbf{p}_{diag}^{-}(i,i)$ equal to 1 for negative $p_i$, and 0 otherwise. $\mathbf{t}=\{{ {t}_{i}}\}_{i=1}^{N}\in {{\mathbb{R}}^{N}}$ denotes timestamps of all $N$ events. $C_{tp}$ denotes output channel. ${{\mathbf{W}}^{+},{{\mathbf{W}}^{-}}\in {{\mathbb{R}}^{1\times {{C}_{tp}}}}}$ and ${{\mathbf{b}}^{+},{{\mathbf{b}}^{-}}\in {{\mathbb{R}}^{{{C}_{tp}}}}}$ are the weight and bias of two different linear layers. Consequently, we concatenate ${\mathbf{F}_{xyp}}$ and ${\mathbf{F}_{tp}}$, and downscale them to a size of C-channel spatiotemporal embedded vectors $\mathbf{F} = {\rm MLP}({\rm Concat}({{\mathbf{F}}_{xyp}},{{\mathbf{F}}_{tp}})) = \left\{ \boldsymbol{f}_{1}, \ldots, \boldsymbol{f}_{N} \right\} \in \mathbb{R}^{N \times \textit{C}}$, where the index of spatiotemporal embedded vector $\boldsymbol{f}_{i}$ correspond one-to-one with that of the event $\boldsymbol e_i$.

\begin{figure*}[t]
\centering
\includegraphics[width=.99\textwidth]{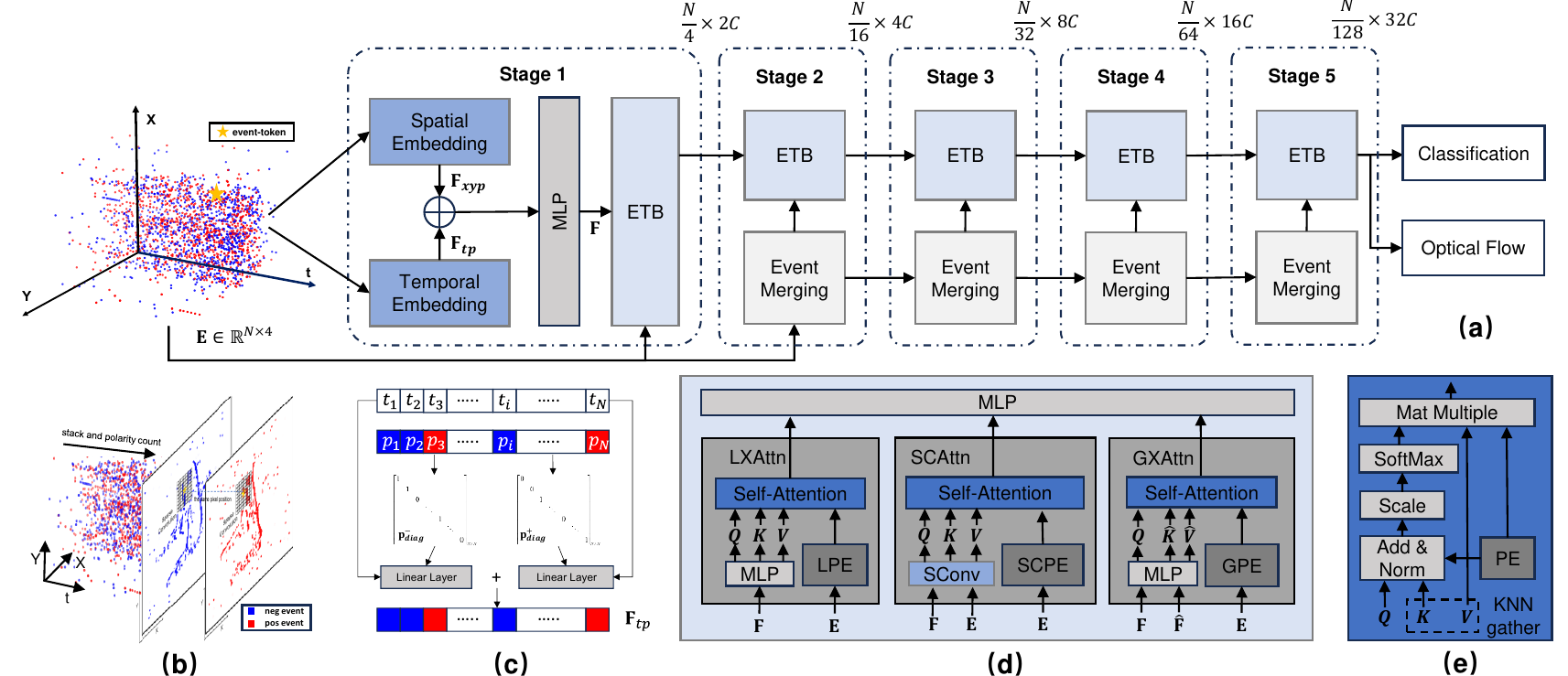}
\label{fig:figure_main}
\caption{\textbf{(a)} Five-stage backbone architecture. Spatial and temporal embedding layer are used at the 1st stage to map input 4-channel attributes of the event-tokens to $C$-channel embedded vectors. From the 2nd to 5th stage, an event merging layer with a farthest event downsampling and an MLP-ReLU-BN are included before each ETB step. For simplicity, we only give the size of features at each stage output. \textbf{(b)} Spatial embedding~\ref{sec:TE}. \textbf{(c)} Temporal embedding~\ref{sec:TE}. \textbf{(d)} Event Transformer Block~\ref{sec:Three-way}. Three-way attention combines different position encoding approaches to construct spatiotemporal correlations between events. \textbf{(e)} Self-attention architecture. PE is position encoding. KNN gather is used for K-Nearest event retrieval.}
\label{fig:figure_main}
\end{figure*}

\subsection{Three-way Attention}
\label{sec:Three-way}
Considering that embedded vectors $\mathbf{F}$ can only extract the spatiotemporal features of individual events and lack the precision to represent temporal sequencing and spatial motion relationships among events, we draw inspiration from the Transformer~\cite{attention_is_all_u_need,PointTransformer} and introduce the Three-way Attention in our Event Transformer Block, depicted in Fig.~\ref{fig:figure_main}(d), to establish spatiotemporal correlations at the event level.

\subsubsection{Local Attention (LXAttn)} The continuity of motion strongly correlates with the local events at the space-time dimension. To construct the spatiotemporal correlation among adjacent events, we employ K-Nearest Neighbors (KNN) to link each event $\boldsymbol{e}_i$ with its $k_l$ temporally closest neighboring events $\mathbf{E}_{KNN}=\{\boldsymbol{e}_j\}_{j=1}^{k_l}$, executing Local Position Encoding (LPE) to obtain LPE vector ${\bf PE}_{ij}^{loc}=\rm MLP(\mathbf{D}_{ij}^{loc})\in \mathbb{R}^{\textit{C}_{l}}$, where $\mathbf{D}_{ij}^{loc}=\sqrt{(x_i-x_j)^2+(y_i-y_j)^2+(t_i-t_j)^2}$ denotes the spatiotemporal distance. 

Then, we derive the $\mathbf{Q},\mathbf{K},\mathbf{V}={\rm{MLP}}_{\boldsymbol{q},\boldsymbol{k},\boldsymbol{v}}(\mathbf{F})\in \mathbb{R}^{ \textit{N}\times \textit{C}_{l}}$, where $\boldsymbol{q}\in \mathbf{Q}$ denotes the query of embedded vector, $\boldsymbol{k}\in \mathbf{K}$ and $\boldsymbol{v}\in \mathbf{V}$ denote the key and value of K-Nearest embedded vector, which can be indexed according to $\mathbf{E}_{KNN}$. Thus, the local spatiotemporal correlations $\textrm{Attn}_i^{loc}$ of event $\boldsymbol{e}_i$ can be written as~\cite{PointTransformer}: 
\begin{equation}
\textrm{Attn}_{i}^{loc}=\sum\limits_{\boldsymbol{e}_{j}} \textrm{softmax}_j \left( \boldsymbol{\textrm{Attn}}_{ij}^{loc} \right) \odot \left( \boldsymbol{v}_{j}+\mathbf{PE}_{ij}^{loc} \right) \label{eq:self-attn}
\end{equation}
where $\textrm{Attn}_{ij}^{loc} = \textrm{MLP} \left( \boldsymbol{q}_i- \boldsymbol{k}_j + \mathbf{PE}_{ij}^{loc} \right)$.

\subsubsection{Sparse Convolution Attention (SCAttn)}
To further clarify the correlation weights of events in the XY dimension, we suggest Sparse Convolution Position Encoding (SCPE). We utilize the ball query algorithm to retrieve $k_{sc}$ events ${{\mathbf{E }}_{win}}=\{{{e}_{j}}\}_{j=1}^{k_{sc}}$ within the local window (e.g., 3×3 in this study) of each event in $\mathbf E$. Thus, the SCPE vector $\mathbf{PE}_{ij}^{sc}=\rm MLP(\mathbf{D}_{ij}^{sc})\in \mathbb{R}^{\textit{C}_{sc}}$, where $\mathbf{D}_{ij}^{sc}=\sqrt{(x_i-x_j)^2+(y_i-y_j)^2}$ denotes the spatial distance. When calculating $\mathbf Q,\mathbf K,\mathbf V$ at spatial scale using sparse convolution, we add the spatiotemporal feature priors to the pixels' feature channels and obtain $\boldsymbol{k}$ and $\boldsymbol{v}$ by retrieving $k_{sc}$ ball query events' indices. The local spatial correlations $\textrm{Attn}_{i}^{sc}$ of $e_i$ can be calculated by modified eq.~\ref{eq:self-attn}.

\subsubsection{Global Attention (GXAttn)}
For high-level vision tasks, the global correlations among events are particularly important. For the purpose of effectively encoding global spatiotemporal information, we use the farthest event down-sampling with down-sample rate $r$ to select the most representative events $\mathbf{\hat{E}}\in \mathbf{E}$. For each $\boldsymbol{e}_g \in \mathbf{\hat{E}}$, its corresponding feature $\boldsymbol{\hat{f}}_g \in \mathbf{\hat{F}}$ is sampled from
$\boldsymbol{\hat{f}}_g = \max_{\boldsymbol{e}_r \in \mathbf{E}_r} \left( {\rm MLP} \left( \textrm{Concat} \left( \boldsymbol{e}_r, \boldsymbol{f}_r \right) \right)  \right)$,
where $\mathbf{E}_r$ is the set of
$\lfloor \frac{1}{\textit{r}} \rfloor$
nearest events of $e_g$, $\boldsymbol{f}_r$ is the embeded vector of $\boldsymbol{e}_r$. Then we derive $\mathbf{Q}={\rm{MLP}}_{\boldsymbol{q}}(\mathbf{F})\in \mathbb{R}^{ \textit{N}\times \textit{C}_{g}}$ and $\mathbf{\hat{K}},\mathbf{\hat{V}}={\rm{MLP}}_{\boldsymbol{\hat k},\boldsymbol{\hat v}}(\mathbf{\hat F})\in \mathbb{R}^{ \textit{N}\times \textit{C}_{g}}$, and the Global Position Encoding (GPE) $\mathbf{PE}_{ig}=\rm{MLP}(\mathbf{D}_{ig})$, where $\mathbf{D}_{ig}$ is the spatiotemporal distance between
each event $\boldsymbol{e}_i$ and its $k_g$ nearest events in $\mathbf{\hat{E}}$. Thus, the global correlations $Attn_{i}^{loc}$ of $e_i$ can be obtained by modified 
eq.~\ref{eq:self-attn}. Finally, $Attn_{i}^{loc}$, $Attn_{i}^{sc}$ and $Attn_{i}^{g}$ are fed into a MLP that outputs the new embedded vectors $\mathbf{F}$ with temporal and spatial correlations between events. The backbone of the proposed model is shown in Fig.~\ref{fig:figure_main}(a).


\section{Experiments}
\subsection{Event-based Object Classification}
\label{sec:exp-classification}
\def\nmnist{N-MNIST}
\def\snmnist{N-M}
\def\ncaltech{N-Caltech101}
\def\sncaltech{N-Cal}
\def\cifar{CIFAR10-DVS}
\def\scifar{CIF10}
\def\ncars{N-Cars}
\def\sncars{N-C}
\def\asl{ASL-DVS}
\def\sasl{ASL}

\newcommand{\Frame}{\textcolor{blue}{$\spadesuit$~}}
\newcommand{\Voxel}{\textcolor{orange}{$\blacksquare$~}}
\newcommand{\Spike}{\textcolor{green}{$\blacklozenge$~}}
\newcommand{\Tensor}{\textcolor{red}{$\bigstar$~}}

\setlength{\tabcolsep}{3pt}
\begin{table}[t]\small
\begin{center}
\caption{Classification Top-1 accuracy. The unit of FLOPs is \SI{}{\giga\relax}. \Frame Frame-based, \Voxel Voxel-based, \Spike Spike-based, \Tensor Token-based. Best result is bolded, the second best is underlined.}
\label{table:classification_results}
\begin{tabular}{lccccccc}
\toprule\noalign{\smallskip}
\textbf{Method}  & \textbf{FLOPs} & \textbf{\snmnist} & \textbf{\sncaltech} & \textbf{\scifar}& \textbf{\sncars} & \textbf{\sasl} \\
\noalign{\smallskip}
\midrule
\noalign{\smallskip}
\Frame EST~\cite{EST}                              & 4.28       & 99.0              & 75.3          & 63.4                  & 91.9           & 97.9            \\
\Frame AMAE~\cite{AMAE}                            & 7.36       & 98.3              & 69.4          & 62.0                  & 93.6           & 98.4            \\
\Frame M-LSTM~\cite{M-LSTM}                        & 4.82       & 98.6              & 73.8          & 63.1                  & 92.7           & 98.0            \\
\Frame MVF-Net~\cite{MVF-Net}                      & 5.62       & 98.1              & 68.7          & 59.9                  & 92.7           & 97.1  \\ 
\Frame EvT~\cite{evt}  & 0.77  &   98.3   &61.3  &59.2  &89.6  &99.9   \\
\Frame DVS-ViT~\cite{dvsvit} & 30.74 & 98.1 & 63.3 & 58.6 & 90.7 & 96.9
\\ \midrule
\Voxel EventNet~\cite{Eventnet}                    & 0.91       & 75.2              & 42.5          & 17.1                  & 75.0           & 83.3            \\

\Voxel RG-CNNs~\cite{asl_dvs}                      & 0.79       & 99.0              & 65.7          & 54.0                  & 91.4           & 90.1            \\
\Voxel EV-VGCNN~\cite{EV-VGCNN}                    & 0.70       & 99.4              & 74.8          & 65.1                  & 95.3           & 98.3            \\
\Voxel VMV-GCN~\cite{VMV-GCN}                   & 1.30       & 99.5              & 77.8          & 66.3                  & 93.2           & 98.9            \\
\Voxel TORE~\cite{TORE}                   & -       & 99.4              & 79.8          & -                  & 94.5           & 99.9            \\
\midrule
\Spike PLIF~\cite{PLIF}                            & 63.21      & 99.6              & -             & 68.5                & -               & -                \\ 
\Spike NDA~\cite{NDA}   & - & -            & 78.2    & 74.5     & 90.1               & -      \\
\Spike VPT-STS ~\cite{VPT-STS}                       & -      & -              & 79.2             & \textbf{82.2}                & \textbf{95.8}              & -                \\ 
\midrule
\Tensor \textbf{Ours}                               & \textbf{0.51} & \textbf{99.9}   & \textbf{81.6}        & \underline{76.4}       & \underline{95.4}       & \textbf{99.9} \\
\bottomrule
\vspace{-1cm}
\end{tabular}
\end{center}
\vspace{0.5cm}
\vspace{-0.5cm}
\end{table}
\setlength{\tabcolsep}{1.4pt}
\noindent \textbf{Datasets and Implementation Details.}
\label{sec:exp-classification-details}
We train our model on the training set of five different datasets (\nmnist~\cite{nMnist}, \ncaltech~\cite{n_caltech101},  \cifar~\cite{cifar10_dvs}, \ncars~\cite{nCars}, \asl~\cite{asl_dvs}) separately and evaluate the Top-1 classification accuracy on the corresponding testing set. We average the output features from the backbone's fifth stage and feed them into a classification head. We set $k_l=k_g=16$ in both LXAttn and GXAttn, and the farthest down-sampling rate $r$ in GXAttn is set to $32, 32, 16, 8, 4$ in each stage separately. In training, $N=2048$ events are randomly sampled from the training set as input without any data augmentation tricks, and all events are input during inference. We train our model from scratch for $200$ epochs by optimizing the cross-entropy loss using the SGD optimizer with momentum $0.9$ and a batch size of $64$. The learning rate starts from $0.01$ and decays to $0.001$ and $0.0001$ at $150$ and $180$ epochs, respectively.

\noindent \textbf{Experimental Results.}
We compare the Top-1 accuracy with other works published in recent years. The models in EST~\cite{EST}, M-LSTM~\cite{M-LSTM}, and MVF-Net~\cite{MVF-Net} were pretrained using ImageNet, and TORE~\cite{TORE} was pretrained using GoogleNet; thus for a fair comparison, we train these networks from scratch without any pretrained model. For PLIF~\cite{PLIF}, NDA~\cite{NDA} and VPT-STS~\cite{VPT-STS}, the original paper divides the dataset into 9:1 for training and testing; therefore, we report the results by retraining their model with 80/20\% train/test split using their official code. We also retrained the EVT~\cite{evt} and DVS-ViT~\cite{dvsvit} with a random sample of $N=2048$ events to compare FLOPs fairly. As shown in Table~\ref{table:classification_results}, we have achieved state-of-the-art classification accuracy on \nmnist, \ncaltech, and \asl. Since we have not found a suitable data augmentation strategy for event-by-event processing, our performance on \cifar, with a limited training set and increased event noise, falls short of \cite{VPT-STS}. Moreover, our model's low computation requirement (only 0.51G FLOPs and 7.39M parameters) stems from fewer parameters and efficient parallel computing. This renders our method well-suited for low-power devices using event cameras, addressing primary computational concerns.
\setlength{\tabcolsep}{4pt}
\begin{table}[]
\begin{center}
\caption{Ablation Study of Temporal Embedding}
\label{table:ablation_temporal}
\resizebox{1\columnwidth}{!}{
\begin{tabular}{lccccc}
\toprule\noalign{\smallskip}
\textbf{Size of Aggregate Range} & \textbf{m=0} & m=3 & m=5 & m=7 & m=9 \\
\noalign{\smallskip}
\midrule
\noalign{\smallskip}
\textbf{Top-1 Accuracy} & \textbf{76.4} & 76.2 & 75.8 & 75.3 & 74.5\\
\bottomrule
\end{tabular}
}
\end{center}
\end{table}
\setlength{\tabcolsep}{1.4pt}

\setlength{\tabcolsep}{1.4pt}

\noindent{\bf Ablation Study.} To verify the effect of temporal fine-grained on event sequence, we conducted ablation experiments at the temporal embedding layer. We aggregate the temporal information of the sequence every $m$ different timestamps and set them uniformly as the average value, thus reducing the temporal granularity. We conducted the ablation study on \cifar, and the results are shown in Table~\ref{table:ablation_temporal}, which reveals that as temporal resolution decreases, the model's classification accuracy also diminishes. This shows the significance of retaining comprehensive temporal information for event sequences. 

\setlength{\tabcolsep}{1.1pt}
\begin{table}\small
\begin{center}
\caption{Optical flow estimation accuracy.}
\label{table:optical_flow_results}
\begin{tabular}{lccccccccccccc}
\toprule\noalign{\smallskip}
\multirow{2}{*}{\textbf{Methods}} &\multirow{2}{*}{\textbf{FLOPs}} & & \multicolumn{2}{c}{\textbf{indoor1}} & & \multicolumn{2}{c}{\textbf{indoor2}} & & \multicolumn{2}{c}{\textbf{indoor3}} & & \multicolumn{2}{c}{\textbf{outdoor1}} \\
 \cmidrule{4-5} \cmidrule{7-8}  \cmidrule{10-11}  \cmidrule{13-14}
 & & & A. & O. & & A. & O. & & A. & O. & & A. & O. \\
\noalign{\smallskip}
\midrule
\noalign{\smallskip}
\Frame EvFlowNet~\cite{ev_flownet}                      & 24.4                & & \underline{1.0}   & 2.2               & & \underline{1.7} & \underline{15} & & 1.5                & 12                &  & \underline{0.5} & \underline{0.2} \\
\Frame FireNet~\cite{firenet}                           & 18.5                & & 1.6               & 4.3               & & 1.9             & 18             & & 1.6                & 13                &  & 0.9 & 2.9 \\ \midrule
\Spike LIF-EFN~\cite{spike-evflownet}                   & 33.4                & & 1.8               & 20                & & 2.6             & 33             & & 2.2                & 24                &  & 0.5 & 0.5 \\
\Spike LIF-FN~\cite{spike-evflownet}                    & \underline{9.1}     & & 1.1               & \textbf{1.8}      & & 1.8             & 15             & & \underline{1.5}    & \underline{9.6}   &  & 0.9 & 1.8 \\ \midrule
\Tensor Ours                                            & \textbf{3.5}        & & \textbf{1.0}      & \underline{2.1}   & & \textbf{1.6}    & \textbf{13}    & & \textbf{1.4}       & \textbf{7.5}      &   & \textbf{0.3} & \textbf{0.1} \\
\bottomrule
\vspace{-0.5cm}
\end{tabular}
\end{center}

\vspace{-0.3cm}
\end{table}

\subsection{Event-based Optical Flow Estimation}
\noindent \textbf{Dataset and Metrics.}
To verify the advantages of event-by-event processing in optical flow estimation, we conducted tests on MVSEC~\cite{mvsec}. Although the indoor data is different again from the outdoor, we only use the outdoor\_day2 sequence for training and the rest for testing following the basic settings in ~\cite{spike-evflownet}.
Average End-point Error (AEE) and Outlier AEE larger than three pixels (OUT) are set as the evaluation metrics, keeping the same as the ~\cite{ev_flownet, spike-evflownet}.
We compute the AEE and OUT for $dt=1$ (about \SI{22}{\milli\second}) since longer interval (e.g.\ $dt=4$) does not cohere with the low-delay characteristic of the event camera. 

\noindent \textbf{Implementation Details.}
We first adopted the U-Net-based architecture to estimate the moving speed of each event. Then, we derive the estimation error by multiplying time $t$ by event speeds. During training, we optimize our model with SGD optimizer for $25000$ iterations, and the learning rate is set to $0.005$ decayed by $0.1$ at $15000$ and $20000$ iterations. The training input is $4096$ random sampled events with a batch size of $12$, whereas all events are fed in at the inference period. We choose Charbonnier loss~\cite{Charbonnier-loss}, a commonly used loss function in this task ~\cite{Charbonnier-loss,ev_flownet}.

\noindent \textbf{Experiment Results.}
Quantitative assessments are presented in Table~\ref{table:optical_flow_results}. All sequences are center-cropped to $256\times256$. For indoor\_flying scenes, we utilize the entire dataset, while for outdoor\_day1, we use only $800$ sequences, as done in~\cite{ev_flownet, spike-evflownet}. Recognizing the substantial impact of varying loss functions on test outcomes, we retrain FireNet~\cite{firenet}, LIF-EvFlowNet~\cite{spike-evflownet}, and LIF-FireNet~\cite{spike-evflownet} using our loss function, as well as EV-FlowNet~\cite{ev_flownet} for equitable comparison. Overall, our model achieves superior performance in terms of AEE across all sequences, consuming only 39\% of the computation cost.

\section{Conclusion}
We present a novel token-based event representation model, treating events as tokens, to embed the spatiotemporal information event by event directly. Drawing inspiration from the transformer architecture, we propose an event transformer backbone that captures temporal and spatial correlations in the event sequence. Our experimental results demonstrate that our proposed method achieves competitive performance while significantly reducing model computation on event-based object classification and optical flow estimation. Our work offers a new approach for processing event data in various vision tasks. 
Moreover, altering the embedding approach could potentially cover frame-based and voxel-grid-based representations.

\section{Acknowledgement}
This work is partially supported by the National Natural Science Foundation of
China under Grant 62022038.  Our code is publicly accessible at {\url{https://github.com/NJUVISION/EventTransformer}}.

\ninept 
\bibliographystyle{IEEEbib-abbrev}
\bibliography{egbib-abbrev}

\end{document}